\documentclass[letterpaper, 10 pt, conference]{ieeeconf}  
\usepackage{amsmath, amsfonts}
\usepackage{algorithmic}
\usepackage{algorithm}
\usepackage{array}
\usepackage[caption=false, font=normalsize, labelfont=sf, textfont=sf]{subfig}
\usepackage{textcomp}
\usepackage{stfloats}
\usepackage{url}
\usepackage{xurl}
\usepackage{verbatim}
\usepackage{graphicx}
\usepackage{cite}
\usepackage{color}
\usepackage{titlecaps}
\usepackage{mathcomp}
\usepackage{bm}

\IEEEoverridecommandlockouts                              

\title{\LARGE \bf
Detachable Wire Drive: \\Reconfigurable Robot Architecture with Shared Actuators
}

\author{Takahiro Hattori$^{1}$, Kento Kawaharazuka$^{1,2}$, Kei Okada $^{1}$ 
  \thanks{$^{1}$ The authors are with the Department of Mechano-Informatics, Graduate School of Information Science and Technology, The University of Tokyo, 7-3-1 Hongo, Bunkyo-ku, Tokyo, 113-8656, Japan.
    {\texttt\small [t-hattori, kawaharazuka, k-okada]@jsk.imi.i.u-tokyo.ac.jp}
  }
	\thanks{$^{2}$ The author is with the AI Center, Graduate School of Information Science and Technology, The University of Tokyo, Japan.
	}
}

\begin{document}
\newcommand{\TODO}[1]{\colorbox{magenta}{{\textbf TODO:}#1}}
\newcommand{\secref}[1]{Section \ref{#1}}
\newcommand{\subsecref}[1]{Subsection \ref{#1}}
\newcommand{\tabref}[1]{{Table \ref{#1}}}
\newcommand{\figref}[1]{{Fig. \ref{#1}}}
\newcommand{\equref}[1]{{Eq. \ref{#1}}}
\newcommand{\algoref}[1]{{Alg. \ref{#1}}}
\newcommand{\enumref}[1]{{\ref{#1}}}
\newcommand{\circletext}[1]{\raisebox{.5pt}{\textcircled{\raisebox{-.5pt} {{\footnotesize #1}}}}}
\newenvironment{ul}{\begin{itemize}}{\end{itemize}}
\newcommand{\li}{\item}

\maketitle
\thispagestyle{empty}
\pagestyle{empty}

\begin{abstract}
Reconfigurable robots offer significant potential for adapting to diverse tasks; however, conventional centralized architectures often require dedicated actuators for each module, leading to substantial increases in overall system weight, volume, and cost. To address these challenges, this paper presents the "Detachable Wire Drive," a reconfigurable robotic system that enables the sharing of heavy and expensive actuators across various morphologies. The core of this system is the "Wire Detach Unit," a mechanism designed to physically split and reconnect wire drive paths, allowing motors to be consolidated into a common base unit. We demonstrate the versatility of this approach by developing a 2-DOF rigid arm, a continuum arm, and two distinct grippers, all of which are interchangeably attached to, and driven by, a single shared actuator set. Experimental results validate the mechanical reliability of the detachment process and the control framework's ability to seamlessly manage transitions between configurations, highlighting a path toward more efficient and multi-functional robotic systems.
\end{abstract}

\section{Introduction}

Various reconfigurable robots have been proposed to adapt to diverse tasks and environments, broadly categorized into decentralized systems (e.g., M-TRAN \cite{mtran}, Room-Bots \cite{roombots}) and centralized systems (e.g., industrial tool changers \cite{electromate_toolchanger,toolchanger_park,toolchanger_das,toolchanger_cheong}, ATRO \cite{atro}, MODMAN \cite{modman}, TRON2 \cite{tron2}). These robots offer advantages such as cost, weight, and volume reduction through common body parts, and simplified maintenance/repair due to reconfigurable structures.
Decentralized systems face challenges including insufficient rigidity, power, and complex software reconstruction due to numerous fine modules. Centralized systems, like industrial tool changers and TRON2 \cite{tron2}, are more practical, achieving rigidity and power through larger modules and clear hierarchical role divisions. However, they suffer from lower component reuse compared to decentralized approaches, limiting cost and weight benefits.
Actuators significantly contribute to a robot's cost and weight (e.g., 30-50\% of weight \cite{weight_ratio_ishida}, with examples like Unitree Go2 motors accounting for 42\% \cite{unitreego2,go2motor}; similarly, 30-50\% of cost \cite{cost_ratio_aparobot}).

\begin{figure}[t]
  \centering
  \includegraphics[width=1.0\linewidth]{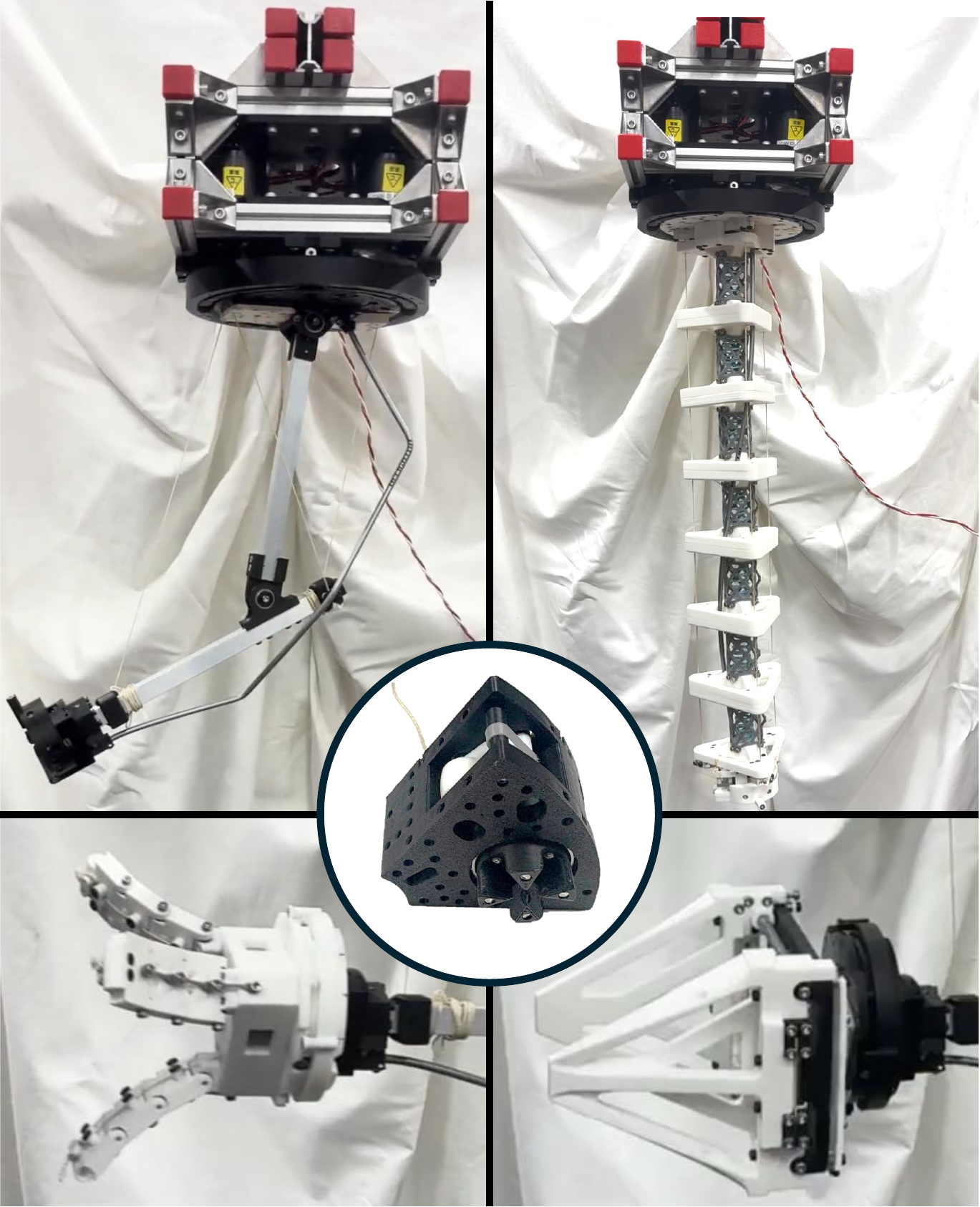}
  \caption{The two types of arms and grippers developed in this study, and the Wire Path Splitting Mechanism (center of the figure) that connects and disconnects them. The two arms are shown attached to the common upper motor-unit.}
  \label{fig:fig1}
\end{figure}

\begin{figure}
  \centering
  \includegraphics[width=1.0\linewidth]{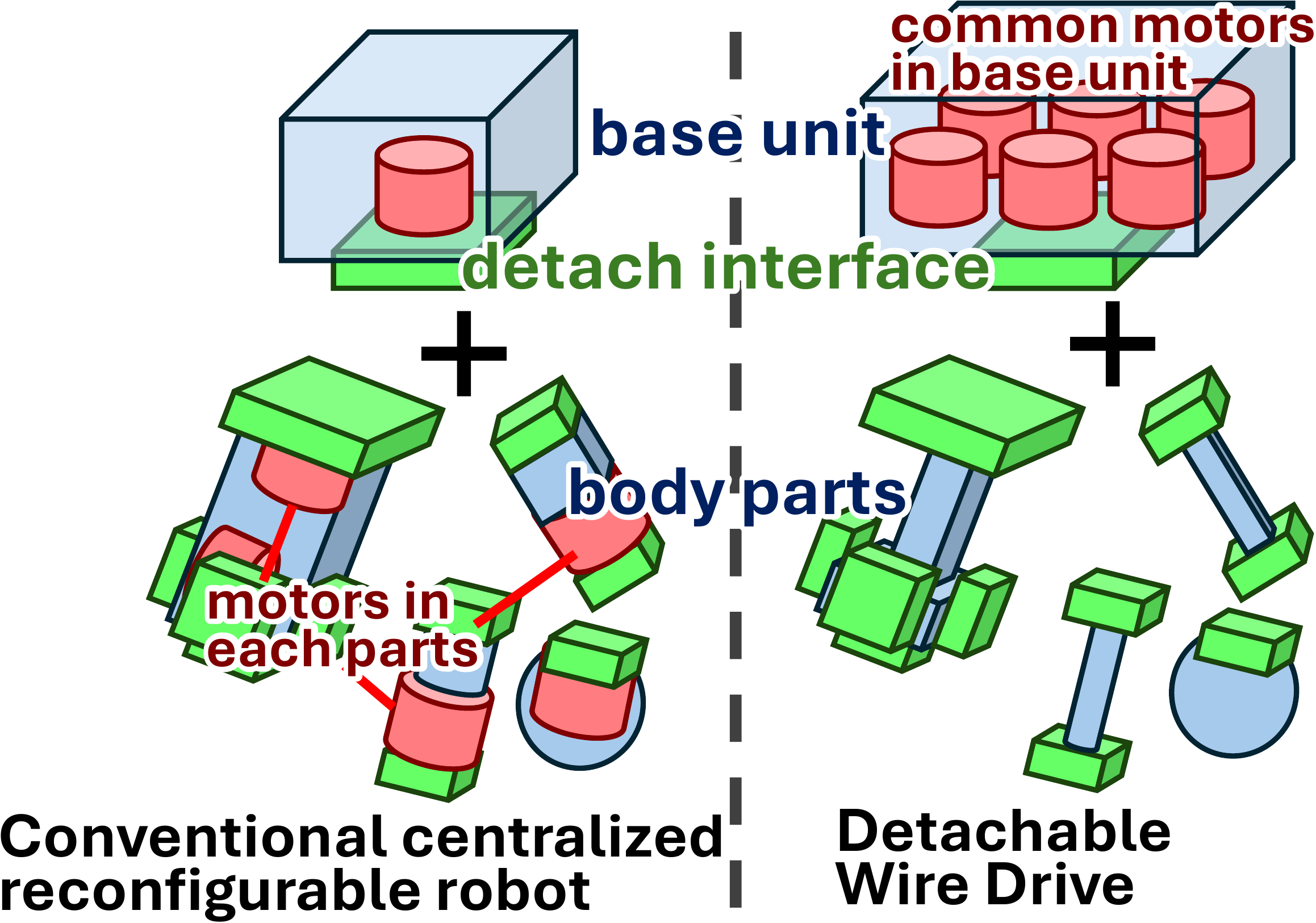}
  \caption{Conceptual diagrams of conventional centralized reconfigurable robots and the proposed reconfigurable wire-driven robot system using a wire path splitting mechanism. Conventional systems require actuators for each module, making it difficult to achieve both morphological versatility and cost/weight savings. In contrast, the proposed system consolidates actuators in a common base unit and transmits power from there, enabling actuator sharing across a wide range of morphologies and achieving both goals. Additionally, using the transmission medium for joint state estimation avoids complex sensor connection reconfigurations during morphological changes.}
  \label{fig:wire_detach_sys_concept_2}
\end{figure}

As shown in \figref{fig:wire_detach_sys_concept_2}, we propose consolidating actuators in a common base unit for centralized reconfigurable robots, transmitting power to drive various body parts. This approach enables actuator sharing across diverse morphologies, achieving versatility and reducing cost/weight. It also simplifies sensor connections by using the transmission medium for joint state estimation. Due to its compactness, high freedom in placement, and sufficient transmission distance, a wire-based mechanism is chosen over hydraulics (leakage) or pneumatics (precision limitations). While wire-driven robots benefit from actuator consolidation at the base to avoid mass/volume constraints (e.g., SAQIEL\cite{SAQIEL}, Super Dragon\cite{super_dragon}), traditional designs with continuous wires are incompatible with detachment. To overcome this, we propose the "Wire Detach Unit," a mechanism allowing wire path splitting with the robot's skeletal structure (see \figref{fig:wire_detach_unit_design}), thereby introducing wire-driven advantages to reconfigurable robots. This study demonstrates the "Detachable Wire Drive," a reconfigurable wire-driven robot system utilizing this unit to achieve broad morphologies with common actuators and electronics. We validate this concept by implementing and controlling two distinct arms (2-DOF rigid, continuum) and two grippers (parallel, multi-fingered) driven by shared actuators, as shown in \figref{fig:fig1}.

The main contributions of this paper are:
\begin{enumerate}
    \item The design of a novel "Wire Detach Unit" that allows for the robust and repeatable connection and disconnection of wire-driven power transmission paths.
    \item The development of the "Detachable Wire Drive," a complete reconfigurable robot architecture that centralizes actuators for shared use across different morphologies.
    \item Experimental validation of the system's reconfigurability and control with two distinct arms and two end-effectors.
\end{enumerate}

\section{Mechanical Design}

\begin{figure}[tbhp]
  \centering
  \includegraphics[width=0.8\linewidth]{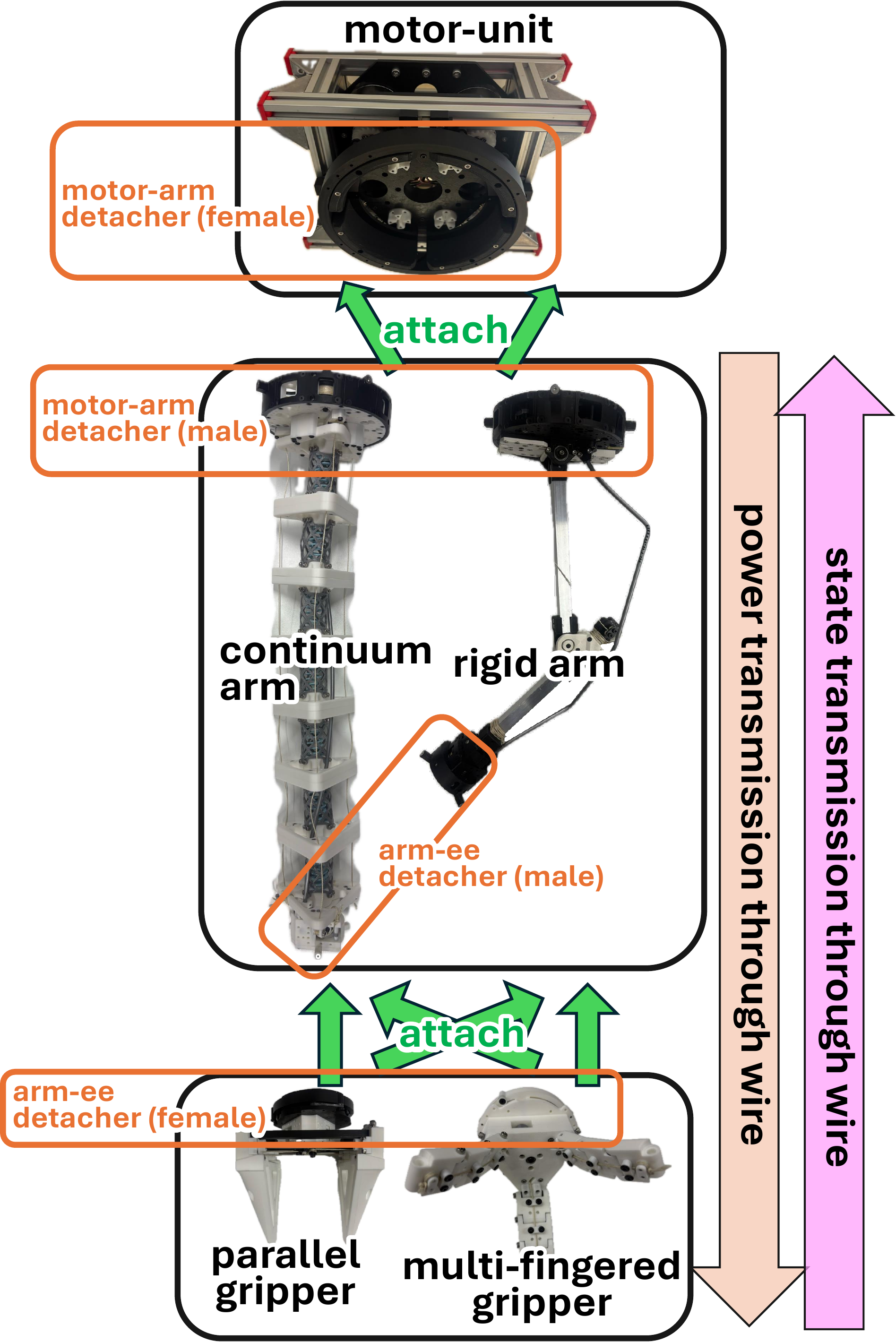}
  \caption{An overview of the reconfigurable wire-driven robot system developed in this study, utilizing the Wire Detach Unit. The system comprises a central motor-unit with motors and electronics, two distinct arm types (a 2-DOF rigid arm and a continuum arm), and two end-effector types (a parallel gripper and a multi-fingered gripper). These components can be interchangeably combined and are driven and controlled by shared actuators.}
  \label{fig:wire_detach_sys_concept_specific}
\end{figure}

The Detachable Wire Drive, illustrated in \figref{fig:wire_detach_sys_concept_specific}, comprises a motor-unit, two distinct arms (2-DOF rigid arm and continuum arm), and two end-effectors (parallel gripper and multi-fingered gripper). Connections between the motor-unit and arms (motor-arm detacher), and arms and end-effectors (arm-ee detacher), are facilitated by the Wire Detach Unit, enabling detachable power and structural coupling. These detachers transmit power via wires from the motor-unit and relay joint state information back through wire displacement.

\subsection{Wire Detach Unit}\label{subsec:wire_detach_unit}
\begin{figure}[tbhp]
  \centering
  \includegraphics[width=1.0\linewidth]{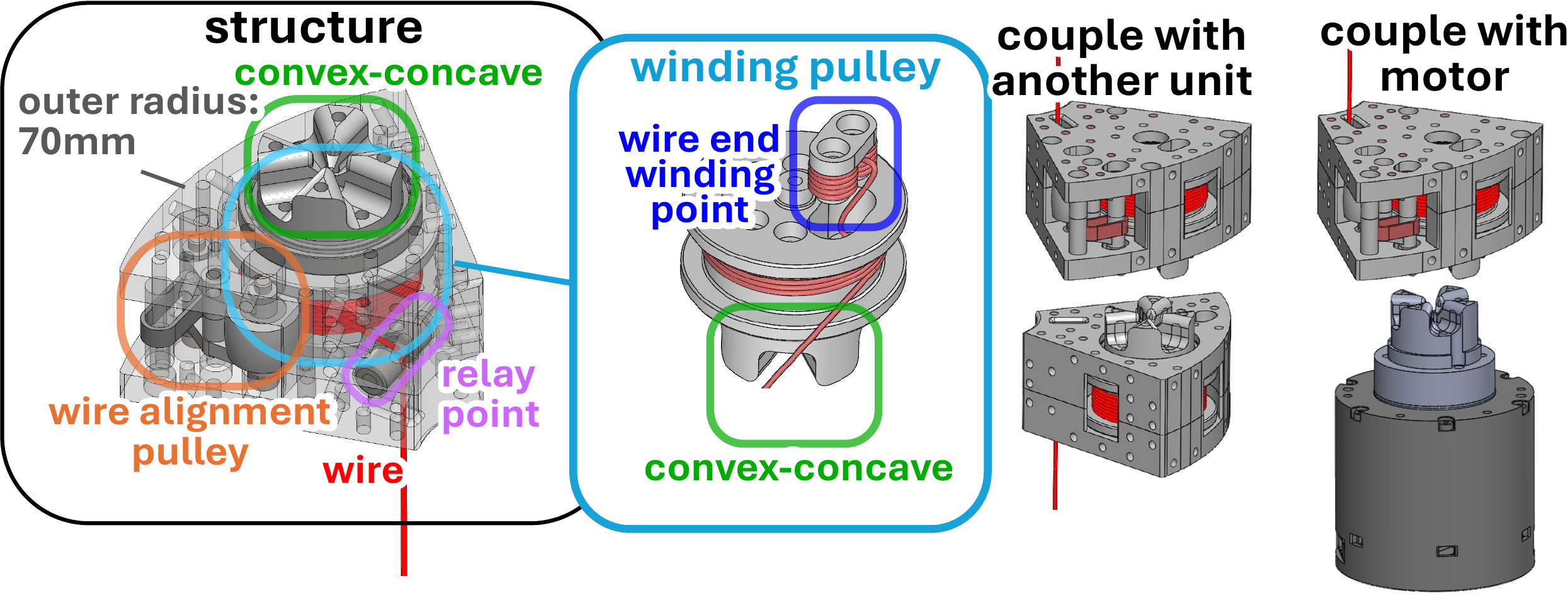}
  \caption{Design of the Wire Detach Unit. It consists of a winding pulley to take up the wire, a wire alignment pulley to ensure stable winding, a relay point to guide the wire, and convex-concave features for torque transmission. The convex-concave features have rounded edges to aid smooth interlocking. The entire unit is housed in a sector-shaped outer shell, allowing up to 6 units to be arranged in a circular pattern for collective attachment and detachment. This mechanism enables the seamless connection and disconnection of both the physical structure and the wire-based power transmission path.}
  \label{fig:wire_detach_unit_design}
\end{figure}

The Wire Detach Unit, detailed in \figref{fig:wire_detach_unit_design}, facilitates robust wire power transmission connection and disconnection. Key components include a Vectran\cite{vectran_about} wire, a winding pulley, a wire alignment pulley, and a relay point. Convex-concave features on the pulley axis enable power engagement. The unit is housed in a sector-shaped outer shell for modular arrangement. Specifications are shown in \tabref{tab:wire_detach_unit_spec}.

\begin{table}
  \centering
  \begin{tabular}{c|c}
    \hline
    \textbf{Specification} & \textbf{Value} \\
    \hline
    wire tensil strength & 1000 N \\
    \hline
    wire diameter & 1 mm \\
    \hline
    wire type & Vectran\cite{vectran_about} \\
    \hline
    winding pulley radius & 15 mm \\
    \hline
    winding pulley width & 10 mm \\
    \hline
    \begin{tabular}{c}
      maximum winding length \\ with same radius 
    \end{tabular} & 942 mm \\
    \hline
    \begin{tabular}{c}
      maximum winding length \\ allowing for radius variation
    \end{tabular}  & 3768 mm \\
    \hline
    weight & 100 g \\
    \hline
    dimensions & 
    \begin{tabular}{c}
      70 mm radius, \\ 60-degree sector
    \end{tabular}\\
    \hline
    maximum unit bundle number & 6 \\
    \hline
    cost per unit & 18 dollars \\
    \hline
  \end{tabular}
  \caption{Specifications of the Wire Detach Unit.}
  \label{tab:wire_detach_unit_spec}
\end{table}

\subsection{Motor-Unit and Motor-Arm Detacher Design}
\begin{figure}[tbhp]
  \centering
  \includegraphics[width=1.0\linewidth]{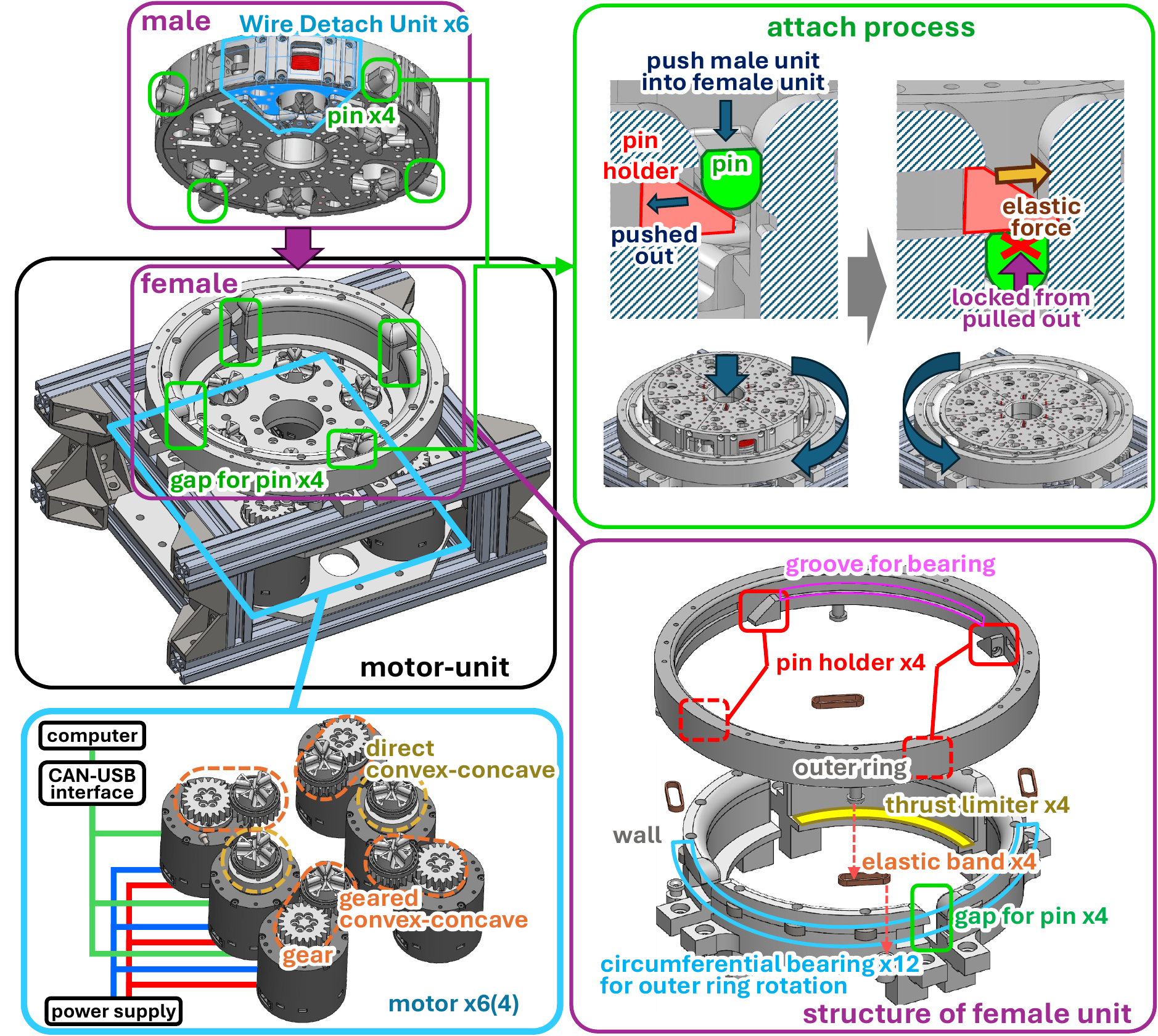}
  \caption{Design of the motor-unit and the motor-arm detacher. The motor-unit (left) houses the motors and features the female side of the detacher. The arm (right) is equipped with the male side, which includes multiple Wire Detach Units. The detacher employs a latch mechanism with pins on the male part and a rotating outer ring with pin holders on the female part, enabling quick and secure attachment and detachment.}
  \label{fig:motor_unit_detach_design}
\end{figure}

The motor-unit, shown in \figref{fig:motor_unit_detach_design}, houses four "RoboStride00"\cite{RS00} QDD motors (10:1 gear ratio, 5 Nm rated, 15 Nm peak torque, 260 rpm speed). With a 16 mm wire winding radius, it provides 312 N continuous tension and 0.44 m/s max speed, meeting arm/gripper requirements. The motor-arm detacher, connecting the motor-unit and arm, uses a male part (on the arm) with six circumferentially arranged Wire Detach Units and four radial pins, engaging with a female part (on the motor-unit). The female part features motor-linked convex-concave elements, a constraining wall, an outer ring with pin holders for secure locking via elastic force, and a thrust limiter. This design allows simple push-to-attach and rotate-and-pull-to-detach operations, ensuring both ease of use and high rigidity.

\subsection{Arm-EE Detacher Design}
\begin{figure}[tbhp]
  \centering
  \includegraphics[width=1.0\linewidth]{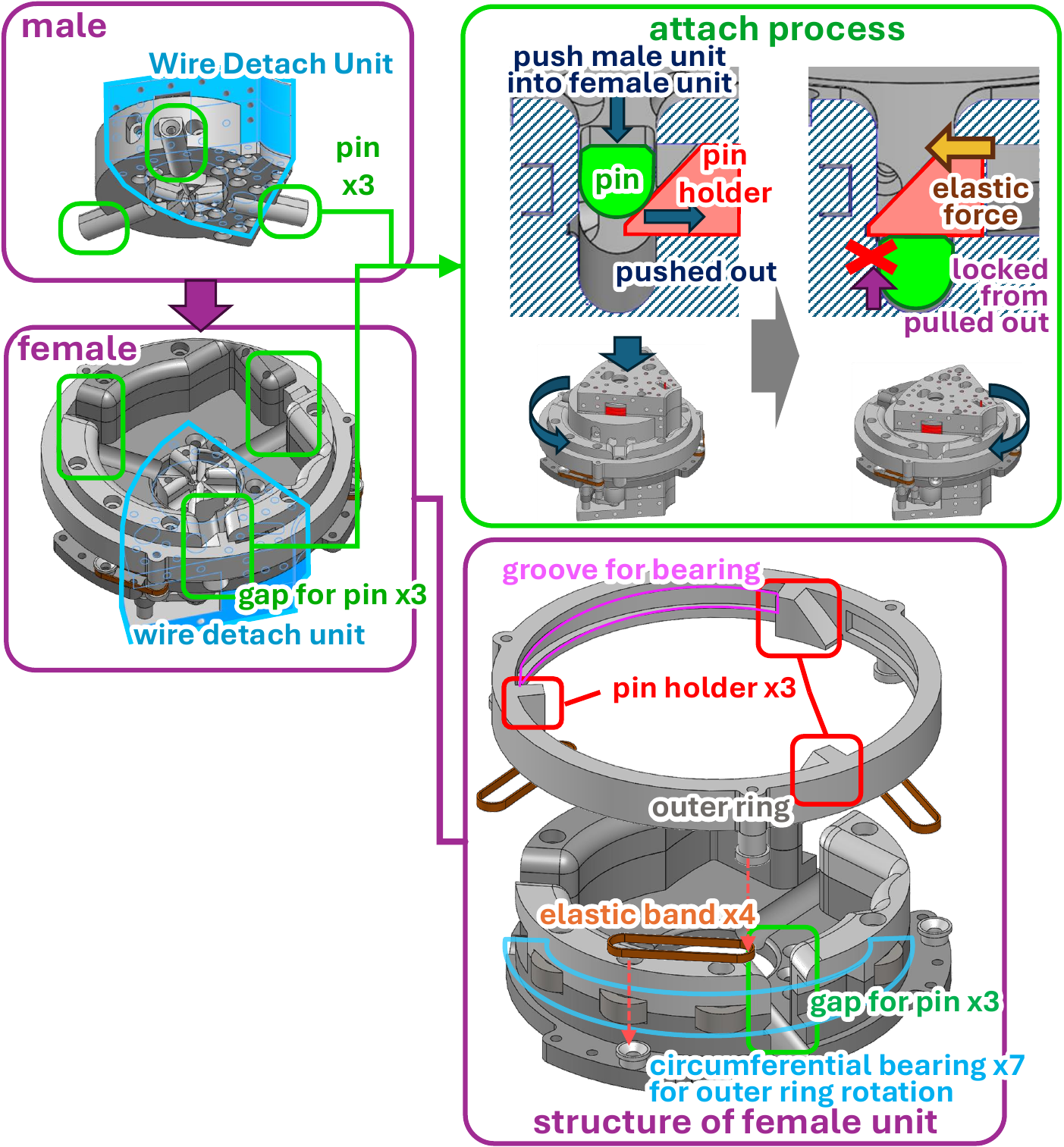}
  \caption{Design of the arm-end-effector detacher. The male part is on the arm side and the female part is on the end-effector side. A single Wire Detach Unit is integrated into each side the mechanism to transmit power to the end-effector. Similar to the motor-arm detacher, it uses a pin-and-locking-ring mechanism for quick attachment and detachment.}
  \label{fig:end_effector_detach}
\end{figure}

The arm-ee detacher, shown in \figref{fig:end_effector_detach}, connects the arm (male part) and end-effector (female part), each containing a Wire Detach Unit for power and structural coupling. Similar to the motor-arm detacher, it employs a pin on the male side and a pin holder with an outer ring on the female side. This enables convenient push-to-attach and rotate-and-pull-to-detach functionality, providing both ease of use and sufficient rigidity.

\subsection{Rigid Arm Design}
\begin{figure}[tbhp]
  \centering
  \includegraphics[width=0.85\linewidth]{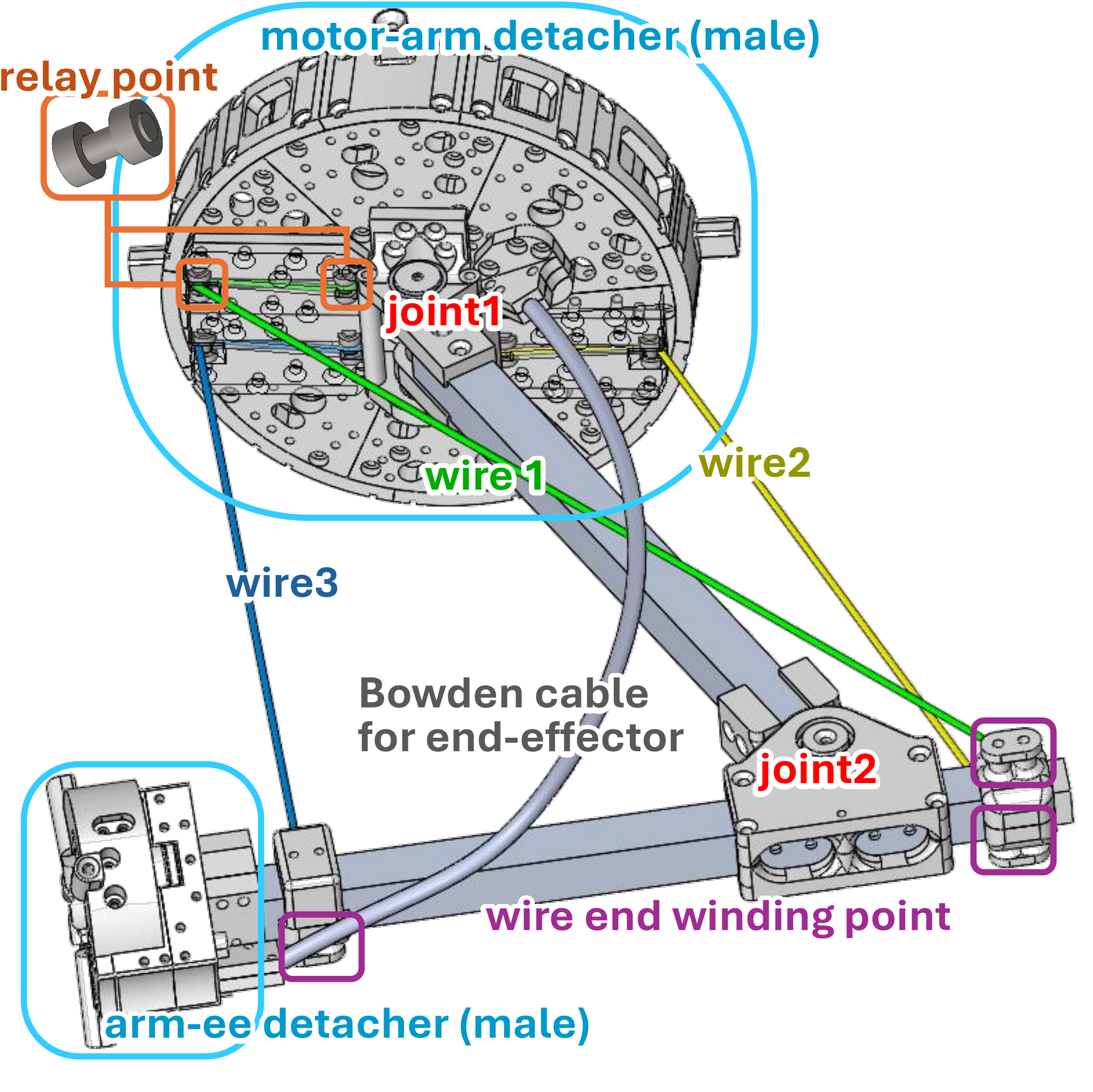}
  \caption{Design of the 2-DOF rigid arm. It features a shoulder joint (joint1) and an elbow joint (joint2). Three wires for arm actuation are routed from the base and attached to the second link. A separate wire for the end-effector is guided through a Bowden cable to the detacher at the arm's tip.}
  \label{fig:rigid_design}
\end{figure}

\figref{fig:rigid_design} illustrates the 2-DOF rigid arm, featuring shoulder and elbow joints. Three actuation wires route from the motor-arm detacher through relay points to the 2nd link. A separate wire for the end-effector passes via a Bowden cable to the arm-ee detacher. Joint 1 and 2 ranges are $-\pi/4$ to $\pi/4$ rad and $-0.6$ to $\pi/4$ rad, respectively. Each joint can exert about $\pm10\sim \pm20$ Nm maximum torque depending on the posture, and has about 10 kg payload, for 450 N maximum wire tension.

\subsection{Continuum Arm Design}
\begin{figure}[tbhp]
  \centering
  \includegraphics[width=1.0\linewidth]{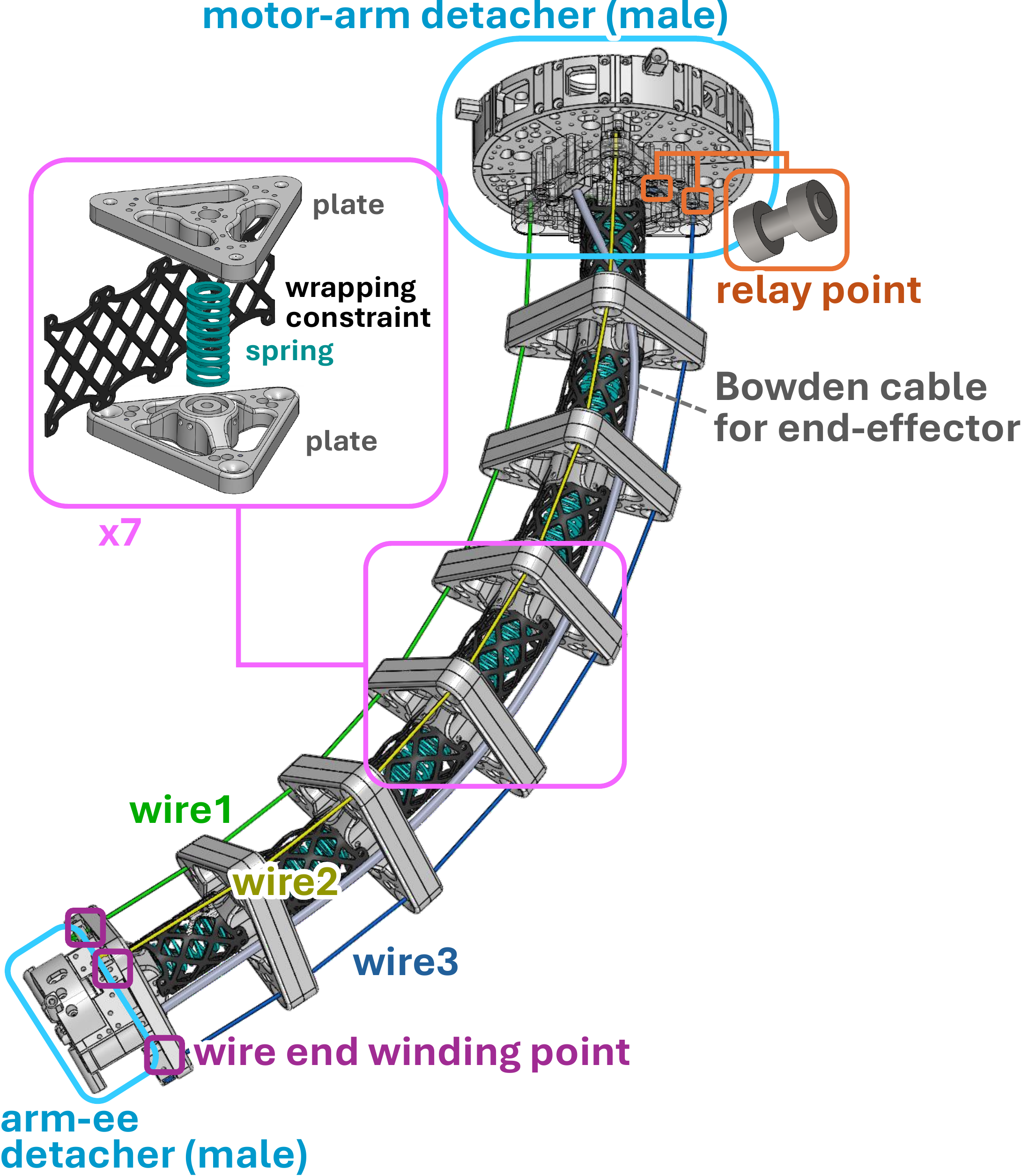}
  \caption{Design of the continuum arm. The arm is constructed from a series of identical units, each composed of a central spring and two triangular plates. Wires for actuation run through the structure to the tip. A flexible sheet wraps around the springs to provide structural integrity and constrain excessive twisting.}
  \label{fig:continuum_design}
\end{figure}

The underactuated continuum arm, shown in \figref{fig:continuum_design}, offers infinite pitch, yaw, and translational DOFs. Three actuation wires from the motor-arm detacher pass through each link's sliding points to the final link's winding point. An end-effector wire uses a Bowden cable to drive the arm-ee detacher. The arm comprises seven identical units, each with a spring, two triangular plates, and a flexible wrapping constraint to prevent twisting and maintain structural integrity. Each joint can exert about $\pm$ 20 Nm maximum bending torque and about 1000 N maximum contracting force for 450 N maximum wire tension.

\subsection{Parallel Gripper Design}
\begin{figure}[tbhp]
  \centering
  \includegraphics[width=0.8\linewidth]{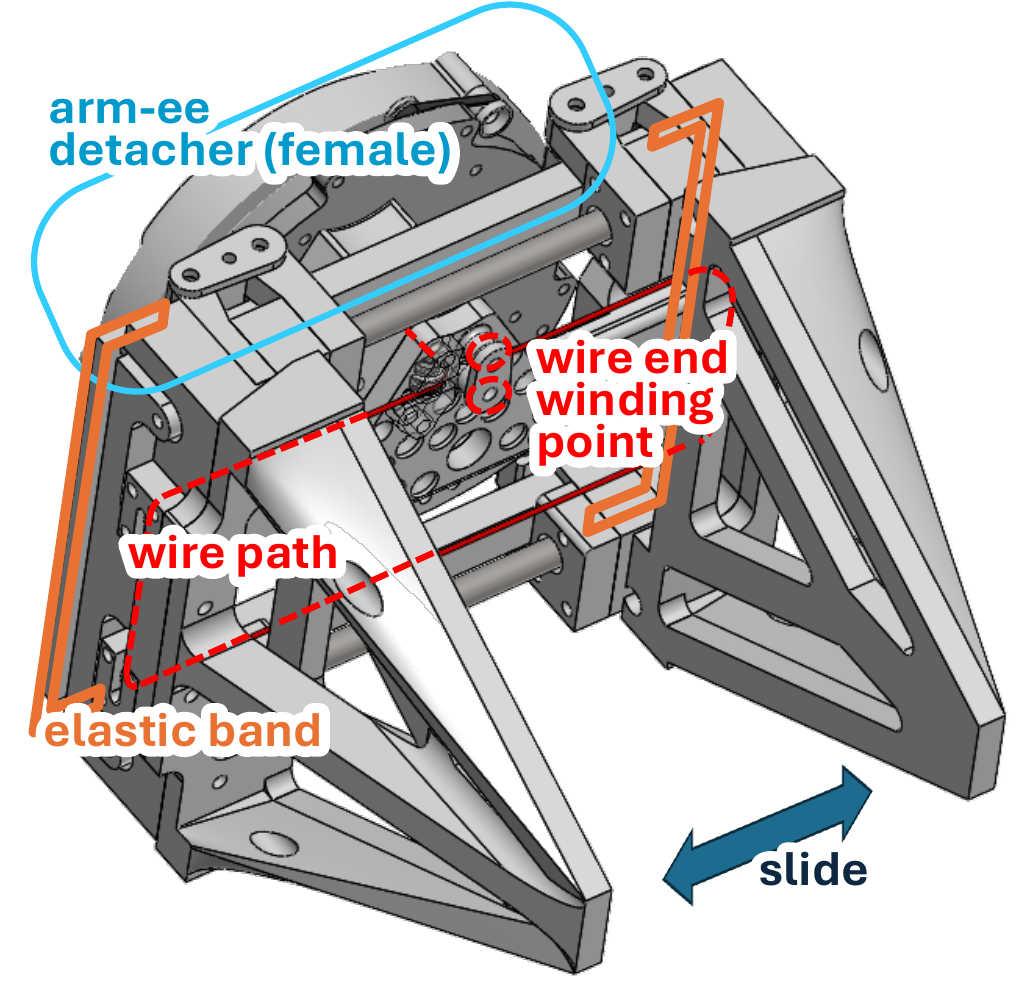}
  \caption{Design of the parallel gripper. It features two jaws that move in parallel. A single wire from the attached Wire Detach Unit drives the closing motion, while an elastic band provides the opening force. Bearings are used at sliding and bending points to ensure smooth operation.}
  \label{fig:parallel_gripper_design}
\end{figure}

\figref{fig:parallel_gripper_design} shows the parallel gripper with two jaws for parallel closing. A wire from the arm-ee detacher drives the closing motion, while an elastic band provides opening force. Bearings ensure smooth movement at wire bending points and linear sliders.

\subsection{Multi-Fingered Gripper Design}
\begin{figure}[tbhp]
  \centering
  \includegraphics[width=0.8\linewidth]{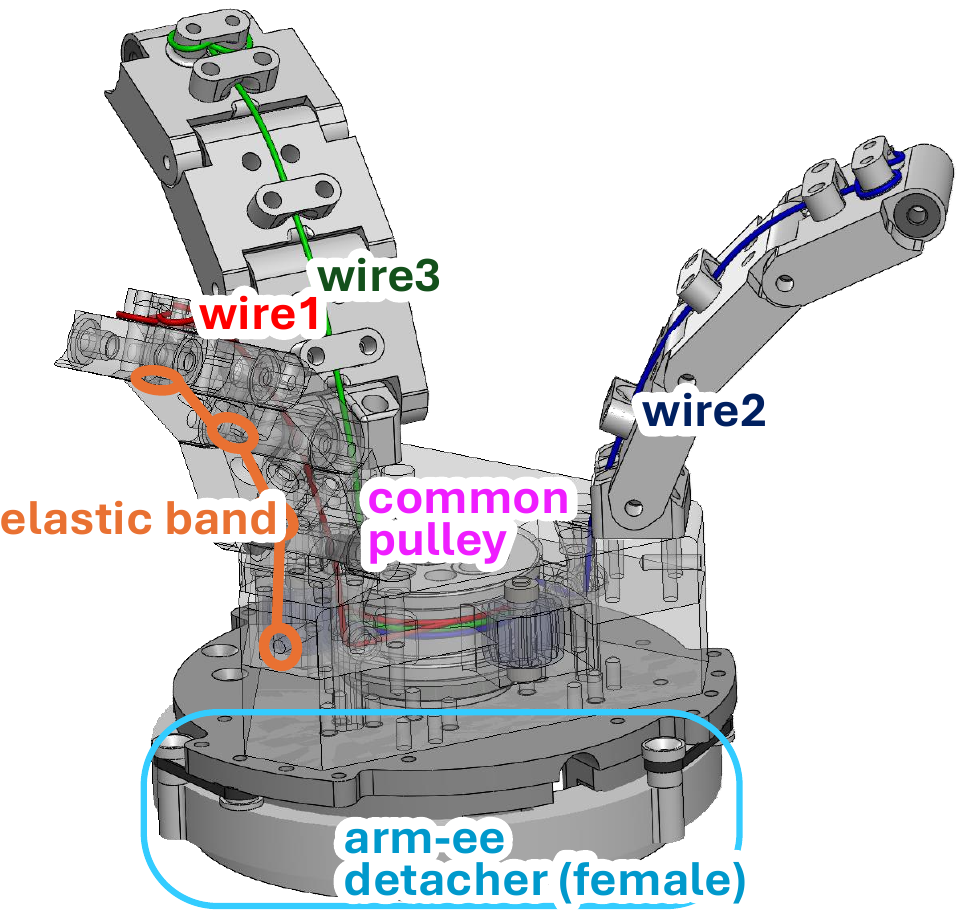}
  \caption{Design of the multi-fingered gripper. It has three underactuated fingers, each with three joints, driven by a single wire for closing and elastic bands for opening. This allows adaptive grasping. A common pulley inside the arm-ee detacher couples the three finger wires, requiring a slight modification to the detacher's Wire Detach Unit.}
  \label{fig:multi_finger_hand}
\end{figure}

The multi-fingered gripper, depicted in \figref{fig:multi_finger_hand}, features three underactuated fingers, each with three joints, driven by a single wire for closing and elastic bands for opening. This allows adaptive grasping. A common pulley inside the arm-ee detacher couples the three finger wires, requiring a slight modification to the detacher's Wire Detach Unit.

\section{Control Method}
This section details the controller for the Detachable Wire Drive, designed for uniform detachment, state estimation, and joint control across diverse body morphologies.

\subsection{State Transition}\label{subsec:state_shift}
\begin{figure}[tbhp]
  \centering
  \includegraphics[width=0.5\linewidth]{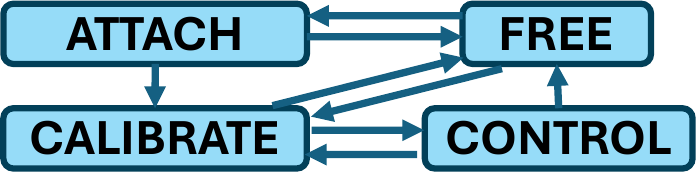}
  \caption{State transition diagram for the controller. The system can be in one of four states: FREE (motors off), ATTACH (active engagement assistance), CALIBRATE (joint angle estimation reset), and CONTROL (closed-loop operation). The diagram shows the permitted transitions between these states, managed according to operator commands and system status.}
  \label{fig:state_shift}
\end{figure}

The controller manages state transitions based on operator input, as shown in \figref{fig:state_shift}. States include: ATTACH, where the motor-side convex-concave features rotate to aid connection by low-gain velocity control (only from FREE); CALIBRATE, for setting a wire winding reference point by applying a specific tension and making the arm settle by gravity to a known reference point or push against a joint limit (from any state); CONTROL, for tracking target joint angles via estimated angle error and controlled wire tension (only from CALIBRATE); and FREE, where tension is zero for safe detachment or recalibration (from any state).

\subsection{Joint Control}\label{subsec:joint_control}
\begin{figure}[tbhp]
  \centering
  \includegraphics[width=1.0\linewidth]{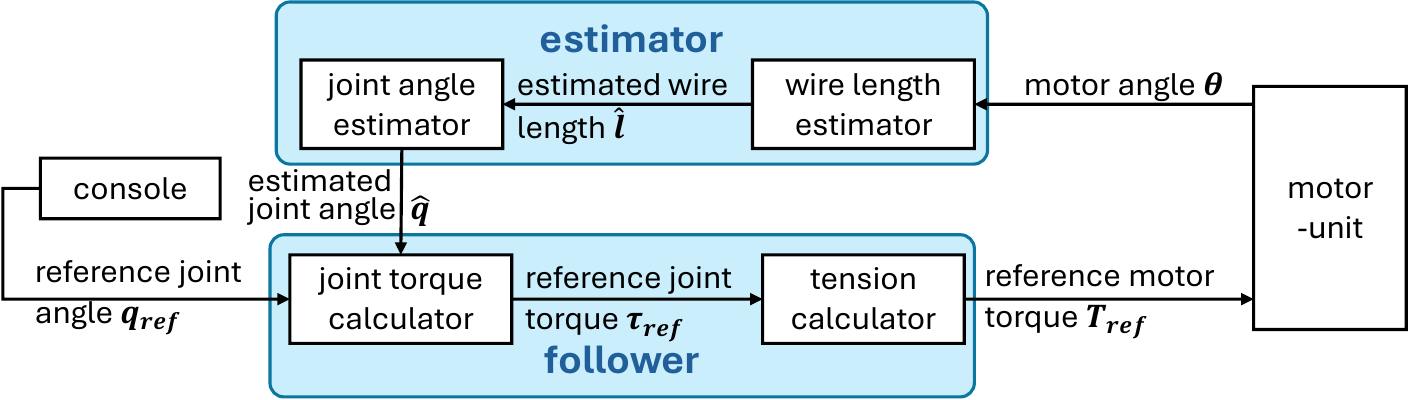}
  \caption{Block diagram of the joint control system. The controller takes a reference joint angle $\bm{q}_{ref}$ and calculates motor torques $\bm{T}_{ref}$. The process involves a Joint Angle Follower, which computes reference wire tensions $\bm{f}_{ref}$ from the joint angle error, and a Joint Angle Estimator, which estimates the current joint angles $\hat{\bm{q}}$ from motor encoder readings $\bm{\theta}$.}
  \label{fig:ctrl_flow}
\end{figure}

For both types of arms, this study uses a method that estimates joint angles from wire displacement, calculates the target joint torque from the error between the estimated and target angles, and then calculates the target tension. The flow of these processes is shown in \figref{fig:ctrl_flow}.

\subsubsection{Joint Angle Estimator}
The relationship between the geometric length vector of the wire inside the body part, $\bm{l}$, and the joint angle vector, $\bm{q}$, is expressed by the function $\bm{H}$ as in \equref{eq:q_l_eq}. Here, $\bm{l}_0$ is a wire length offset determined by where the base end of the wire is located.
\begin{equation}
  \bm{l} - \bm{l}_0 = \bm{H}(\bm{q}).
  \label{eq:q_l_eq}
\end{equation}
The geometric length vector of the wire is expressed using the motor's winding angle vector $\bm{\theta}$ and the wire's winding radius $r$ as in \equref{eq:l_theta_eq}.
\begin{equation}
  \bm{l} = -r\bm{\theta}.
  \label{eq:l_theta_eq}
\end{equation}

As mentioned in \subsecref{subsec:state_shift}, the reference point for the wire winding amount is set after aligning to a specific joint angle $\bm{q}_0$ in the CALIBRATE state. Therefore, $\bm{l}_0$ is expressed using the motor winding amount $\bm{\theta}_0$ in the CALIBRATE state as in \equref{eq:l0_calib_eq}.

\begin{equation}
  \bm{l}_0 = - r\bm{\theta}_0 - \bm{H}(\bm{q}_0).
  \label{eq:l0_calib_eq}
\end{equation}

Using $\bm{l}$ and $\bm{l}_0$ obtained from \equref{eq:l_theta_eq} and \equref{eq:l0_calib_eq}, the estimated joint angle $\hat{\bm{q}}$ is found by solving an optimization problem as in \equref{eq:l_to_q_opt}. Here, $\bm{q}_{min}$ and $\bm{q}_{max}$ are the physical constraints of the joint angles. The $\hat{\bm{q}}$ from the previous step is used as the initial value.
\begin{equation}
    \begin{aligned}  
    &\min_{\hat{\bm{q}}}  \qquad \left \| \bm{l} - \bm{l}_0 - \bm{H}(\hat{\bm{q}}) \right \|^2
    \\ 
    &\text{s.t.} \qquad \bm{q}_{min} \leq \hat{\bm{q}} \leq \bm{q}_{max}
    \end{aligned}
  \label{eq:l_to_q_opt}
\end{equation}

For underactuated arms like the continuum arm, $\hat{\bm{q}}$ cannot be uniquely determined from the wire length. To stabilize the estimation, we assume that the joint angles for the repeating units for pitch, yaw, and translation are the same. Specifically, in the robot model definition including the wires, we define master joints and slave joints that take the same angle, and only the master joints are treated as elements of the joint angle vector $\bm{q}$ in \equref{eq:q_l_eq}, \equref{eq:l0_calib_eq}, and \equref{eq:l_to_q_opt}. By considering underactuation and elastic forces in the model definition format and applying the same estimation and tracking framework to multiple body parts, we can uniformly control a wide range of morphologies.

\subsubsection{Joint Angle Follower}

To make the estimated joint angle $\hat{\bm{q}}$ follow $\bm{q}_{ref}$, we use a method\cite{torque_control} that calculates the target joint torque from joint error and then calculates the target tension from that torque.

In this study, the joint torque vector $\bm{\tau}_{ref}$ is the sum of the torque from PID control, gravity compensation torque, and elastic force compensation torque from springs, etc., as shown in \equref{eq:calc_tauref}.

\begin{equation}
  \begin{array}{l}
    \bm{\tau}_{ref} \\
    = \left( K_p (\bm{q}_{ref} - \hat{\bm{q}}) + K_d (\dot{\bm{q}}_{ref} - \dot{\hat{\bm{q}}}) + K_i \int (\bm{q}_{ref} - \hat{\bm{q}}) dt \right)\\ 
    \qquad + \bm{\tau}_{gravity} + K_{elastic}\hat{\bm{q}}
  \end{array}
  \label{eq:calc_tauref}
\end{equation}

The conversion from $\bm{\tau}_{ref}$ to the target tension vector $\bm{f}_{ref}$ is as follows.

First, the conversion from the tension vector $\bm{f}$ to the target joint torque vector $\bm{\tau}$ is expressed as in \equref{eq:f_tau_eq}.
\begin{equation}
  \bm{\tau} = -\bm{G}^T \bm{f}.
  \label{eq:f_tau_eq}
\end{equation}

Here, $\bm{G}$ is called the Tendon Jacobian and is defined as in \equref{eq:tendon_jacobian_eq}. Geometrically, $\bm{G}_{i,j}$ represents the moment arm and direction of wire $j$ with respect to joint $i$.
\begin{equation}
  \bm{G} = \frac{\partial \bm{H}(\bm{q})}{\partial \bm{q}}.
  \label{eq:tendon_jacobian_eq}
\end{equation}

From \equref{eq:f_tau_eq}, the conversion from the target joint torque vector $\bm{\tau}_{ref}$ to the target tension vector $\bm{f}_{ref}$ is expressed as an optimization problem as in \equref{eq:tau_to_f_opt}. Here, $\bm{f}_{min}$ and $\bm{f}_{max}$ are the minimum and maximum constraints on the wire tension.
\begin{equation}
    \begin{aligned}  
    &\min_{\bm{f}_{ref}}  \qquad \left \| \bm{f}_{ref} \right \|^2 +  \left( \bm{\tau}_{ref} + \bm{G}^T \bm{f}_{ref} \right)^T \bm{\Lambda} \left( \bm{\tau}_{ref} + \bm{G}^T \bm{f}_{ref} \right)
    \\ 
    &\text{s.t.} \qquad \bm{f}_{min} \leq \bm{f}_{ref} \leq \bm{f}_{max}
    \end{aligned}
  \label{eq:tau_to_f_opt}
\end{equation}

The calculated target tension vector $\bm{f}_{ref}$ is multiplied by the pulley radius $r$ and sent to the motor as the target motor torque $\bm{T}_{ref}$, thereby controlling the system to follow the target joint angle $\bm{q}_{ref}$.
\section{Experiment}
Several experiments were conducted to evaluate the effectiveness of the Detachable Wire Drive.
\subsection{Wire Detach Unit Evaluation Experiment}
\begin{figure}[tbhp]
  \centering
  \includegraphics[width=1.0\linewidth]{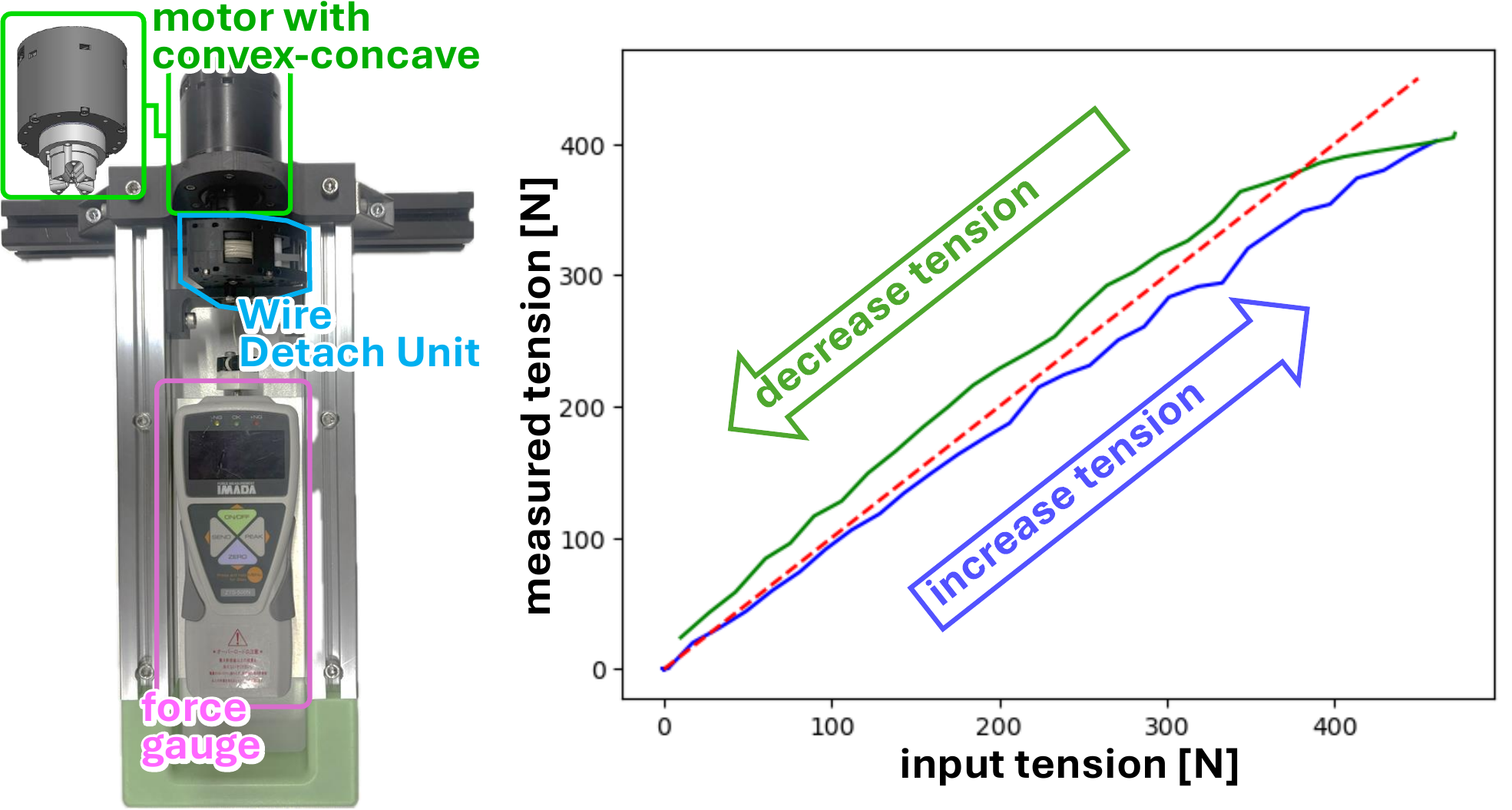}
  \caption{Results of the strength and friction evaluation experiment for the Wire Detach Unit. Input tension was cycled from 0 to 450 N and back to 0 over 60 seconds. The deviation between input and output tension averaged approximately $\pm$16 N, which is sufficiently small compared to the ideal direct correspondence (red dotted line). The unit successfully withstood 450 N of tension without failure.}
  \label{fig:unit_exp}
\end{figure}

The Wire Detach Unit's strength and friction were evaluated by engaging it with a motor and connecting its wire to a force gauge. Tension was cycled from 0 to 450 N over 60 seconds. Results in \figref{fig:unit_exp} show an average $\pm$16 N friction deviation, deemed sufficiently small. The unit withstood 450 N tension without failure.

\subsection{Motor-Unit Arm Sequential Detachment and Control Experiment}
\begin{figure*}[tbhp]
  \centering
  \includegraphics[width=1.0\linewidth]{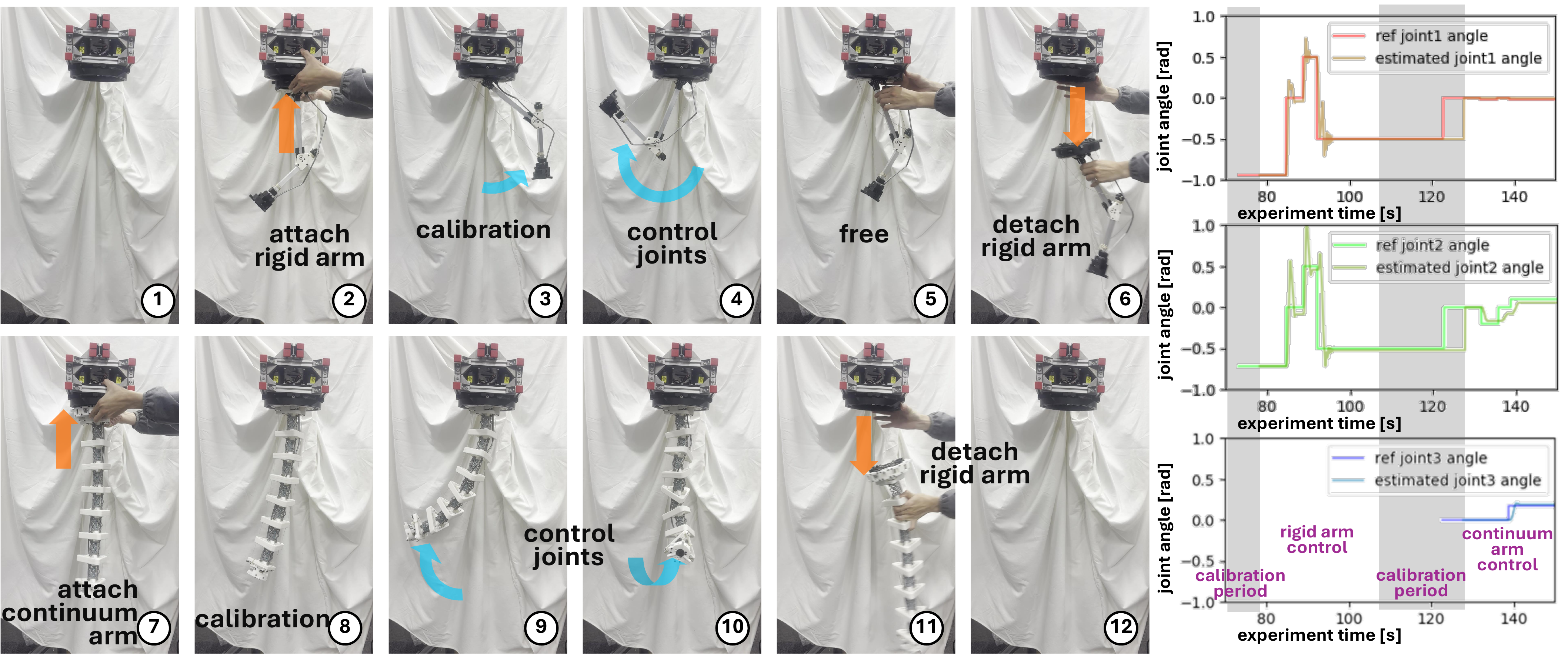}
  \caption{Experiment showing the sequential attachment, control, and detachment of two different arms (rigid and continuum) from the motor-unit. The graphs on the right show the target and estimated joint angle transitions for both arms during control. In both cases, the estimated joint angles successfully track the target joint angles.}
  \label{fig:motor_unit_detach_exp}
\end{figure*}

To validate the motor-arm detacher and seamless mechanical/control switching, we sequentially attached, controlled, and detached two arm types (rigid and continuum) from the motor-unit, as shown in \figref{fig:motor_unit_detach_exp}. This involved ATTACH (pushing arm onto motor-unit with rotating convex-concave features), CALIBRATE (setting angle limits or gravity-settled posture), CONTROL (tracking random target angles), and FREE (detaching by rotating outer ring). Both arms successfully tracked target angles in CONTROL, confirming seamless operation. Attachment required about 10 seconds to wait for the convex-concave features to rotate and engage, while detachment took about 1 second to rotate the outer ring and disengage. This demonstrates the system's ability to quickly switch between different arm configurations.

\subsection{End-Effector Detachment and Manipulation Experiment}
\begin{figure*}[tbhp]
  \centering
  \includegraphics[width=1.0\linewidth]{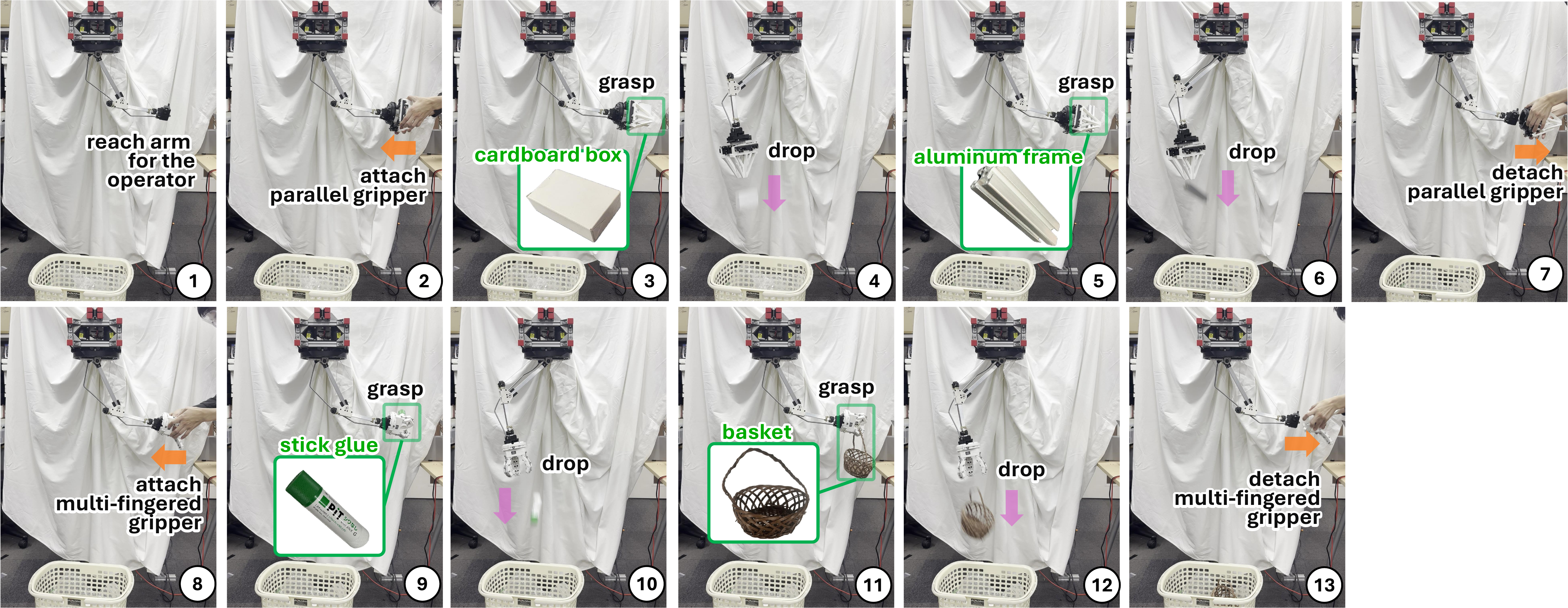}
  \caption{Experiment showing the sequential attachment, control, and detachment of two different grippers on the rigid arm. Both grippers demonstrate the ability to properly grasp and manipulate objects according to their shape and size.}
  \label{fig:hand_detach_exp}
\end{figure*}

The arm-ee detacher's functionality and object-adaptive grasping were evaluated by sequentially attaching, controlling, and detaching two grippers (parallel and multi-fingered) to the rigid arm, as shown in \figref{fig:hand_detach_exp}. Grippers were attached by pushing and detached by rotating the outer ring. Each gripper successfully grasped cardboard boxes and aluminum frames, and stick glue and baskets, demonstrating effective manipulation in CONTROL state via discrete wire tension. Attachment and detachment required about 1 second each, allowing for quick switching between end-effectors.

\subsection{Rigid Arm Tracking Performance Evaluation Experiment}
\begin{figure}[tbhp]
  \centering
  \includegraphics[width=1.0\linewidth]{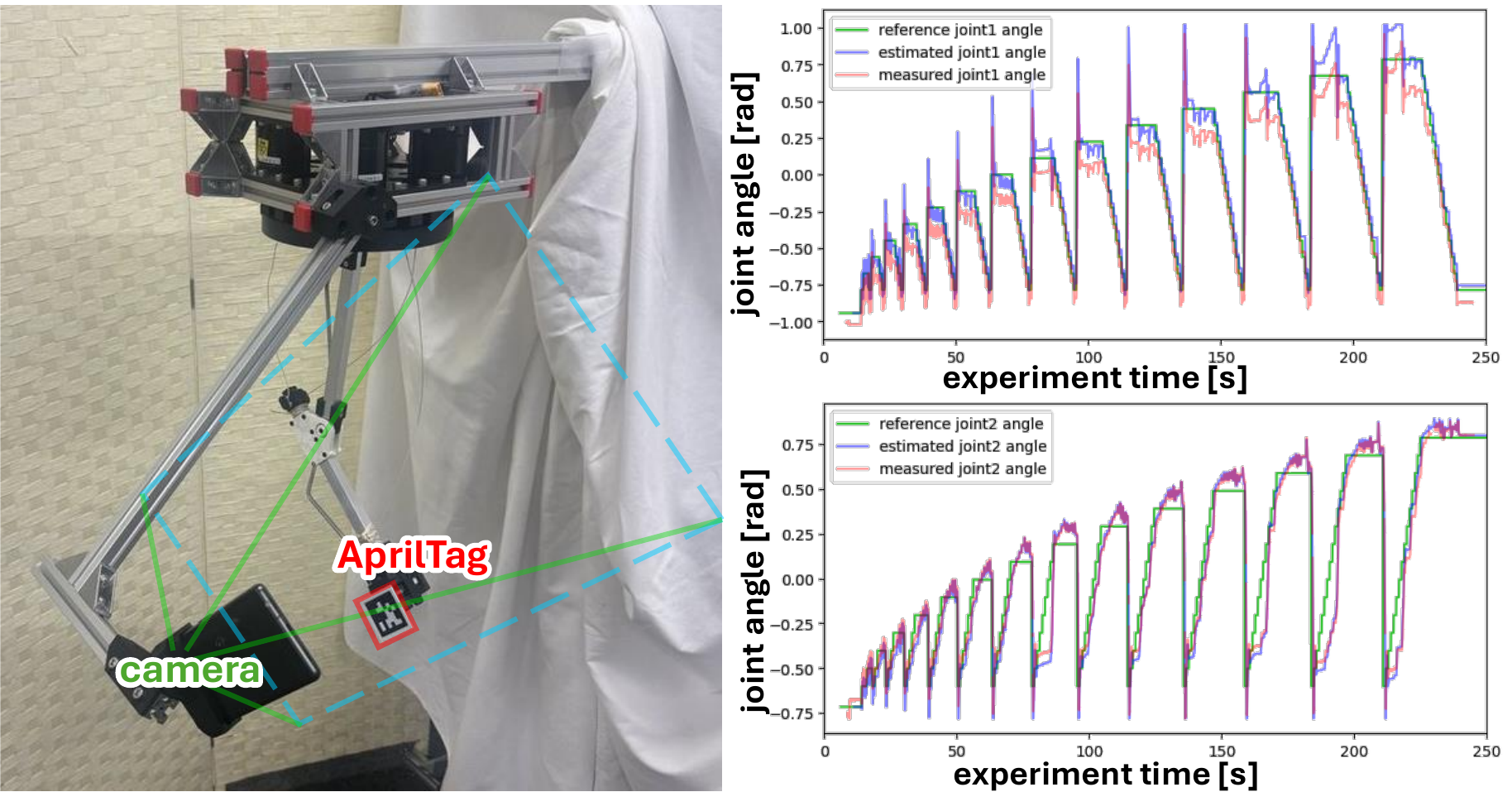}
  \caption{Transition of target joint angle $\bm{q}_{ref}$, estimated joint angle $\hat{\bm{q}}$, and actual joint angle $\bm{q}$ in the joint control state of the rigid arm. It can be seen that $\hat{\bm{q}}$ generally follows $\bm{q}$, and $\bm{\hat{q}}$ generally follows $\bm{q}_{ref}$. }
  \label{fig:follow_exp}
\end{figure}

The joint controller's performance was evaluated for the rigid arm using vision-based AprilTag measurements due to the absence of joint encoders and the continuum arm's hyper-redundancy. Joint 1 and 2 ranges were divided into 15 steps each, perturbing 225 target joint angles. \figref{fig:follow_exp} and \tabref{tab:follow_exp} show that estimated joint angles accurately tracked actual and target angles, confirming sufficient estimation and tracking performance.

\begin{table}[tbhp]
  \centering
  \begin{tabular}{c|c}
    \hline
    \textbf{Comparison} & \textbf{RMSE [rad]} \\
    \hline
    $\bm{q}$ vs. $\hat{\bm{q}}$ & 0.087 \\
    $\bm{q}_{ref}$ vs. $\hat{\bm{q}}$ & 0.145 \\
    $\bm{q}_{ref}$ vs. $\bm{q}$ & 0.177 \\
    \hline
  \end{tabular}
  \caption{RMSE evaluation of the error between the target joint angle $\bm{q}_{ref}$, estimated joint angle $\hat{\bm{q}}$, and actual joint angle $\bm{q}$ in the joint control state of the rigid arm. The results show that the joint angles can be estimated with sufficient accuracy from the wire length, and that there is sufficient tracking of the target joint angles.}
  \label{tab:follow_exp}
\end{table}

\section{Discussion}
The results of the Wire Detach Unit evaluation experiment showed that the unit can withstand tension of about 450 N and transmit wire tension without significant friction.

The results of the motor-unit arm sequential detachment and control experiment demonstrated that both the rigid arm and the continuum arm can be quickly attached to and detached from the motor-unit using the motor-arm detacher while maintaining rigidity and stability when attached. It was also shown that after attachment, by transitioning to the CONTROL state, both the rigid arm and the continuum arm can be made to follow target joint angles with a common controller. Furthermore, it confirmed that this series of operations can be controlled continuously.

The results of the end-effector detachment and manipulation experiment showed that both the parallel gripper and the multi-fingered gripper can be attached to and detached from the rigid arm without any rigidity and transmission problems using the arm-ee detacher. It also confirmed that the attached grippers could properly grasp their respective target objects, and that manipulation such as grasp and drop could be performed in combination with arm movements. The end-effector attachment was faster than the motor-arm attachment, which is likely because the arm-ee detacher has only one transmission path, which required far less time for the convex-concave features to rotate and engage compared to the motor-arm detacher with six transmission paths.

Finally, the results of the rigid arm tracking performance evaluation experiment showed that the joint angles of the rigid arm could be estimated with sufficient accuracy from the wire length, and that there was sufficient tracking of the target joint angles. The slight vibration and overshoot observed may be due to factors other than insufficient gain tuning, such as: 1. Modeling errors in estimating joint angles from wire length. There are elements that are difficult to model completely, such as modeling the winding at the wire end as a point, or the slight changes in length and winding radius as the wire sags or thins depending on tension. 2. Control delay. The $\bm{H}(\bm{q})$ in \subsecref{subsec:joint_control} involves geometric calculations based on the robot model, and calling such a function about three times on average in the optimization process results in a somewhat slow control cycle of about 50 Hz. A higher frequency is likely needed to prevent vibration in a transmission medium like wire, which can be taut or slack. Therefore, it is necessary to consider faster estimation methods to improve control performance.

\section{Conclusion}
In this paper, we proposed the "Detachable Wire Drive," a reconfigurable wire-driven robot system designed to share actuators across different morphologies, thereby reducing overall system cost and weight. The core of this system is the "Wire Detach Unit," a novel mechanism that enables the structural and power-transmission connections and disconnections. We demonstrated the feasibility of this concept by developing a system comprising a central motor-unit, two distinct arms (a 2-DOF rigid arm and a continuum arm), and two different end-effectors (a parallel gripper and a multi-fingered gripper).

Experiments confirmed that the Wire Detach Unit can transmit substantial force with low friction. We successfully demonstrated the quick and reliable attachment and detachment of different arms and end-effectors, and showed that a unified control framework can estimate joint states from wire displacement and effectively control these varied configurations. The system proved capable of performing manipulation tasks by swapping end-effectors to suit different objects. While the tracking performance was generally good, some oscillations suggest that faster estimation methods and more refined kinematic models could further improve performance.
This work validates the principle of sharing actuators in a centralized, reconfigurable wire-driven system, paving the way for more cost-effective, lightweight, and versatile robotic platforms.



{
  \bibliographystyle{IEEEtran}
  \bibliography{bib}
}

\end{document}